\newif\ifcameraready
\camerareadyfalse
\newif\ifpreprint
\preprinttrue   
\newif\ifnamedbuild
\ifcameraready\namedbuildtrue\fi
\ifpreprint\namedbuildtrue\fi

\documentclass{article}

\usepackage{microtype}
\usepackage{graphicx}
\usepackage{subcaption}
\usepackage{booktabs}
\usepackage{hyperref}

\ifcameraready
  \usepackage[accepted]{icml2026}        
\else
  \ifpreprint
    \usepackage[preprint]{icml2026}      
  \else
    \usepackage{icml2026}                
  \fi
\fi

\ifpreprint
  \makeatletter
  \renewcommand{\ICML@preprint}{\textit{Preprint. A shorter version was accepted (non-archival) at the ICML 2026 Workshop on Efficient Multimodal Question Answering (EMM-QA).}}
  \makeatother
\fi

\usepackage{amsmath,amssymb,amsthm,mathtools}
\usepackage{algorithm}
\usepackage{algorithmic}
\usepackage[capitalize,noabbrev]{cleveref}

\graphicspath{{figures/}}
\newcommand{\MedVRAG}{{\upshape\textsc{MedVRAG}}}

\icmltitlerunning{Iterative Multimodal RAG for Medical QA}

\begin{document}

\twocolumn[
  \icmltitle{Iterative Multimodal Retrieval-Augmented Generation\\for Medical Question Answering}

  \icmlsetsymbol{equal}{*}

  \ifnamedbuild
    \icmlsetsymbol{corr}{*}
    \begin{icmlauthorlist}
      \icmlauthor{Xupeng Chen}{corr,nyu}
      \icmlauthor{Binbin Shi}{thu}
      \icmlauthor{Chenqian Le}{nyu}
      \icmlauthor{Jiaqi Zhang}{ind}
      \icmlauthor{Kewen Wang}{cas}
      \icmlauthor{Ran Gong}{nyu}
      \icmlauthor{Jinhan Zhang}{nyu}
      \icmlauthor{Chihang Wang}{nyu}
    \end{icmlauthorlist}
    \icmlaffiliation{nyu}{New York University, New York, USA}
    \icmlaffiliation{thu}{Tsinghua University, Beijing, China}
    \icmlaffiliation{ind}{Independent Researcher}
    \icmlaffiliation{cas}{Chinese Academy of Sciences, Beijing, China}
    \icmlcorrespondingauthor{Xupeng Chen}{xc1490@nyu.edu}
  \else
    \begin{icmlauthorlist}
      \icmlauthor{Anonymous Authors}{anon}
    \end{icmlauthorlist}
    \icmlaffiliation{anon}{Anonymous Institution}
    \icmlcorrespondingauthor{Anonymous Authors}{anonymous@example.com}
  \fi

  \icmlkeywords{medical QA, retrieval-augmented generation, multimodal retrieval, vision-language models, iterative reasoning}

  \vskip 0.3in
]

\printAffiliationsAndNotice{}

\begin{abstract}
Text-based medical retrieval-augmented generation (RAG) pipelines lose table structure, figures, and page layout when biomedical documents are parsed into chunks, although much of the evidence in a biomedical document lives in exactly those elements. We ask whether retrieving document \emph{pages as images} and letting a vision-language model (VLM) reason over them iteratively closes this gap. \MedVRAG{} indexes $\sim$350K PubMed Central (PMC) page images with a multi-vector visual retriever (ColQwen2.5), narrows the candidates with a sharded LLM filter over page summaries, and lets a VLM refine its query across up to three rounds while accumulating findings in a structured memory bank. On MedQA, MedMCQA, PubMedQA, and MMLU-Med, \MedVRAG{} reaches 78.6\% average accuracy, $+5.8$ points over the same Qwen2.5-VL-32B backbone without retrieval. A component-at-a-time ablation attributes $+1.0$ to page-image retrieval over text-chunk retrieval, $+1.5$ to iteration, and $+1.0$ to the memory bank; a manual analysis of MedQA questions that the full system answers correctly and the text-chunk baseline does not finds that most of them depend on tables, figures, or layout. Stage-1 retrieval over the full corpus runs in under 30\,ms through a coarse-to-fine index, and a single reasoning round costs $\sim$15.9\,s on 4$\times$A100; on MedQA, most of the errors that iteration recovers are recovered by the second round. We report these results from a single seeded run and discuss what evidence is still needed to make the modality claim statistically firm.
\end{abstract}

\section{Introduction}
\label{sec:intro}

Large language models (LLMs) answer medical licensing-exam questions at expert level~\citep{nori2023capabilities,singhal2025toward}, but they do so from static parametric knowledge and remain prone to hallucination~\citep{ji2023survey}; any use in clinical decision support~\citep{singhal2023large} would require answers grounded in inspectable evidence. Retrieval-augmented generation (RAG)~\citep{lewis2020retrieval} supplies that grounding from an external corpus, and in medicine MedRAG~\citep{xiong2024benchmarking}, Self-BioRAG~\citep{jeong2024improving}, i-MedRAG~\citep{xiong2025improving}, and RAG2~\citep{sohn2025rag2} have all shown substantial gains on standard benchmarks.

These systems share one assumption: retrieval over text. Biomedical PDFs are OCR'd or parsed into chunks, and only the extracted text is indexed. Linearization discards, or garbles, the dosage tables, diagnostic flowcharts, reference ranges, and anatomical diagrams through which much biomedical evidence is actually communicated. A second limitation is single-pass retrieval: many clinical questions are multi-hop, with mechanism, contraindications, and monitoring guidelines living on different pages, and a single retrieval pass can miss evidence for such questions.

We propose \textbf{\MedVRAG{}} (Medical Visual RAG), which addresses both limitations. \emph{Page-level multimodal retrieval} indexes $\sim$350K original PMC document pages with a vision-language embedding model rather than text chunks, so tables and figures are retrieved and read in their native form. \emph{Iterative reasoning with memory} lets a VLM decide whether the retrieved evidence suffices and, if not, issue a refined query while accumulating per-round findings in a structured memory bank, for up to three rounds. General-domain page-image RAG~\citep{faysse2024colpali,yu2025visrag,cho2024m3docrag} targets document visual question answering, and medical RAG systems that iterate~\citep{xiong2025improving} operate on text; to our knowledge, \MedVRAG{} is among the first systems to combine page-image retrieval over the biomedical literature with iterative VLM reasoning and to evaluate the combination on standard medical multiple-choice benchmarks.

\textbf{Contributions.} (1)~Under a matched pipeline (same backbone, filter, prompts, decoding, and evaluation protocol), retrieving PMC pages as images improves over retrieving OCR'd text chunks on all four benchmarks, and a manual analysis of 100 MedQA questions that the full system wins over the text-chunk baseline attributes most of them to tables, figures, or layout cues (\cref{sec:analysis}). (2)~Iterative query refinement with a structured memory bank adds a further gain on top of single-round page retrieval; on MedQA, most of the errors that iteration recovers are recovered by the second round, which makes the round cap a useful deployment knob (\cref{sec:iteration}). (3)~A retrieval design that scales multi-vector page-image search to $\sim$350K pages at $\sim$30\,ms per query, a four-class error taxonomy of the remaining failures, and prompts, schema, decoding settings, and baseline configuration inlined in the appendix.

\section{Related Work}
\label{sec:related}

\textbf{Medical RAG.} MedRAG~\citep{xiong2024benchmarking} benchmarks RAG for medical QA across five corpora and four retrievers under a common protocol (MIRAGE), which we adopt. Self-BioRAG~\citep{jeong2024improving} learns when to retrieve and adds self-reflection; i-MedRAG~\citep{xiong2025improving} issues iterative follow-up queries; RAG2~\citep{sohn2025rag2} filters retrieved passages with rationale-guided relevance. All four retrieve text chunks. \MedVRAG{} keeps the MIRAGE protocol and the iterate-then-answer control flow of i-MedRAG, but replaces the text index with page images and the text reader with a VLM, and adds an explicit memory bank across rounds.

\textbf{Iterative and adaptive retrieval.} Outside medicine, IRCoT~\citep{trivedi2023ircot} interleaves retrieval with chain-of-thought steps, FLARE~\citep{jiang2023flare} retrieves when the generator's confidence drops, Self-RAG~\citep{asai2024selfrag} learns reflection tokens that decide whether to retrieve and assess the relevance and support of the retrieved evidence, and Search-o1~\citep{li2025searcho1} lets a reasoning model issue search calls mid-trace. These methods study generator-controlled retrieval on multi-step QA and long-form generation. \MedVRAG{} uses the same idea with a VLM as the controller, and the controller's state between rounds is an explicit, inspectable record of findings rather than the generation prefix.

\textbf{Page-image retrieval and RAG.} ColPali~\citep{faysse2024colpali} introduced multi-vector late-interaction retrieval over page images with a vision-language backbone; ColQwen2.5 is the community variant that swaps in Qwen2.5-VL, and DSE~\citep{ma2024dse} shows that single-vector screenshot embeddings are competitive retrievers at Wikipedia scale. VisRAG~\citep{yu2025visrag} closes the loop with a VLM generator over retrieved page images, M3DocRAG~\citep{cho2024m3docrag} extends this to multi-page, multi-document corpora, and VDocRAG~\citep{tanaka2025vdocrag} pre-trains a visual-document retriever and trains a generator over the retrieved pages. These RAG systems are evaluated on document VQA over general-domain or enterprise documents and retrieve once per question; none refines the query or keeps state across retrieval rounds. \MedVRAG{} applies the same retrieval family to the biomedical literature, evaluates on knowledge-intensive medical QA rather than document VQA, and adds the iterate-with-memory loop. Multi-vector search over 350K pages is an engineering constraint of our setting that motivates the coarse-to-fine index of \cref{sec:stage1}.

\textbf{Multimodal medical RAG and VLMs.} MMed-RAG~\citep{xia2025mmedrag} and RULE~\citep{xia2024rule} retrieve for an input that is a \emph{medical image} plus a question (radiographs, pathology) and calibrate how much a medical VLM should rely on the retrieved context; the query modality is the image and the corpus is text or image--report pairs. \MedVRAG{} is the converse: the query is text and the \emph{corpus} is visual. LLaVA-Med~\citep{li2023llavamed} and Med-Flamingo~\citep{moor2023medflamingo} are biomedical VLMs trained for visual QA over medical images; we instead rely on the document and layout understanding of a general VLM, Qwen2.5-VL~\citep{bai2025qwen25vl}, without further fine-tuning.

\section{Method}
\label{sec:method}

\begin{figure*}[t]
  \centering
  \includegraphics[width=\textwidth]{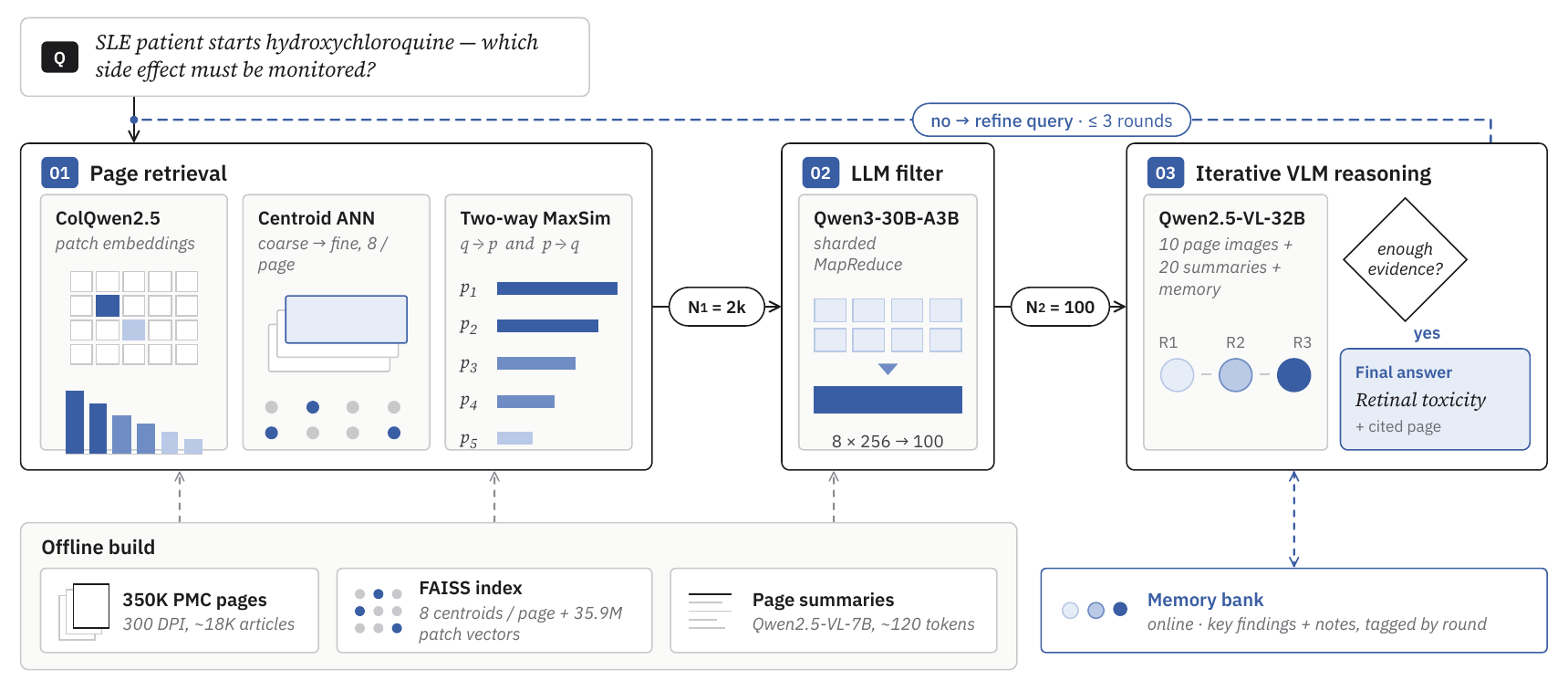}
  \caption{\MedVRAG{} pipeline. A medical question is encoded by ColQwen2.5 and matched against a FAISS index over $\sim$350K PMC page images (Stage~1, top-$N_1{=}2{,}000$); an LLM ranks the candidates by their page summaries with a sharded MapReduce filter (Stage~2, top-$N_2{=}100$); a VLM reasons over the 10 highest-ranked page images plus the top-20 summaries and the memory bank, and either answers or issues a refined query that triggers a new retrieval round ($\le 3$ rounds). The corpus, index, and page summaries are built offline; the memory bank is updated after every non-final round.}
  \label{fig:arch}
\end{figure*}

\MedVRAG{} has four components (\cref{fig:arch}): an offline page-image index over the PMC corpus (\cref{sec:kb}); a two-stage retrieval pipeline of embedding search followed by LLM filtering (\cref{sec:stage1,sec:stage2}); an iterative VLM reasoner (\cref{sec:reasoning}); and a memory bank that carries findings across rounds (\cref{sec:memory}). \Cref{alg:loop} gives the inference loop.

\subsection{Knowledge Base and Page Embeddings}
\label{sec:kb}

We render each article in the PMC Open Access Subset page-by-page at 300\,DPI and drop near-duplicate pages (ColQwen2.5 cosine similarity $>0.97$). The final corpus is $\sim$350K pages from $\sim$18K articles. Articles overlapping with the source articles of MedQA, MedMCQA, and PubMedQA are excluded by DOI/PMID; MMLU-Med has no source DOIs, so we additionally verify that no retained page contains verbatim question text.

Each page is encoded with the ColPali multi-vector page-image retriever~\citep{faysse2024colpali}, specifically the ColQwen2.5 community variant that swaps PaliGemma for a Qwen2.5-VL backbone, into $n$ patch-level vectors ($\sim$103 per page on average), which we PCA-reduce to $d{=}128$ (from 768) to bound the index size. A short text summary ($\sim$120 tokens) of every page is also generated offline by Qwen2.5-VL-7B for use in Stage~2.

\subsection{Stage 1: Coarse-to-Fine Embedding Retrieval}
\label{sec:stage1}

Given a question $q$ encoded into $m$ token vectors $\{q_i\}$, and a page $p$ with patch vectors $\{v_j\}$, we score the pair with a two-way late-interaction sum
\begin{equation*}
S(q,p) = \tfrac{1}{m}\sum_i \max_j q_i^\top v_j \;+\; \tfrac{1}{n}\sum_j \max_i q_i^\top v_j ,
\end{equation*}
i.e., the standard query-to-page MaxSim plus its page-to-query counterpart, so that pages whose patches are not explained by the query are penalized. Computing this exactly against 350K pages is $O(m\!\cdot\!n\!\cdot\!N)$ per query. We instead $k$-means-cluster each page's patches offline into $C{=}8$ centroids and store the centroids, each with a pointer to its page, in a FAISS ANN index; online, the $m$ query vectors are searched against the centroid index to obtain a shortlist of top-$R$ candidate pages, and the exact two-way score is computed only for those $R$ pages with their full patch sets. The cost drops to $O(m\!\cdot\!C\!\cdot\!N + m\!\cdot\!n\!\cdot\!R)$ and runs in under 30\,ms per query (\cref{tab:latency}). Stage~1 returns the top $N_1{=}2{,}000$ pages.

\subsection{Stage 2: Sharded LLM Filter}
\label{sec:stage2}

The $N_1$ candidates are re-ranked by an LLM (Qwen3-30B-A3B) reading their page summaries. Since $N_1\times\bar\ell_{\rm sum}\approx 240$K tokens exceeds the 32K context, we use a sharded MapReduce: 8 parallel map calls each rank a shard of $B{=}256$ summaries ($\sim$31K tokens) and emit their top-25, and one reduce call ranks the surviving 200 into the final $N_2{=}100$ pages. The filter prompt is in \cref{app:prompts}.

\subsection{Iterative VLM Reasoning}
\label{sec:reasoning}

In each round $t\in\{1,2,3\}$ the VLM (Qwen2.5-VL-32B-Instruct, hereafter Qwen2.5-VL-32B) receives the question and options, the \textbf{10 highest-ranked page images} from the $N_2{=}100$ filtered set at native dynamic resolution, the top-20 page summaries, and the memory bank. It either emits an answer inside \texttt{<answer>} tags or a refined query inside \texttt{<query\_update>} tags together with notes on what it has found so far; a refined query triggers a fresh Stage-1 and Stage-2 retrieval in the next round. At the third round the prompt requires an answer. Answers are extracted by a fixed regex; outputs without an \texttt{<answer>} tag count as incorrect.

\begin{algorithm}[t]
  \caption{\MedVRAG{} inference loop}
  \label{alg:loop}
  \begin{algorithmic}[1]
    \STATE $q_1 \leftarrow$ question; \; $M \leftarrow \emptyset$ \hfill $\triangleright$ memory bank
    \FOR{$t = 1, \dots, K{=}3$}
      \STATE $\mathcal{P}_1 \leftarrow \textsc{Stage1}(q_t)$ \hfill $\triangleright$ top-$N_1{=}2{,}000$ pages
      \STATE $\mathcal{P}_2 \leftarrow \textsc{Stage2}(q_t,\mathcal{P}_1)$ \hfill $\triangleright$ top-$N_2{=}100$ pages
      \STATE $o \leftarrow \textsc{VLM}\big(\text{question, options},\ \text{images}(\mathcal{P}_2[{:}10]),$
      \STATE \hspace{3.2em} $\text{summaries}(\mathcal{P}_2[{:}20]),\ M\big)$
      \IF{$o$ contains \texttt{<answer>} \OR $t = K$ \hfill $\triangleright$ round $K$ forces an answer}
        \STATE \textbf{return} answer extracted from $o$
      \ENDIF
      \STATE $q_{t+1} \leftarrow$ refined query in $o$
      \STATE $M \leftarrow M \cup \{\text{notes in } o, \text{ tagged } [\text{Round } t]\}$
    \ENDFOR
  \end{algorithmic}
\end{algorithm}

\subsection{Memory Bank}
\label{sec:memory}

The memory bank is a JSON object with three fields: \texttt{iteration}, the round counter; \texttt{key\_findings}, a list of facts extracted so far, each prefixed with \texttt{[Round k]} so that later rounds can attribute evidence to the round that found it; and \texttt{reasoning\_history}, a list of per-round notes. After every non-final round the VLM's notes are appended to the bank, and the bank's findings are rendered as text and inserted into the next round's prompt. Its purpose is twofold: it prevents findings from being lost when the retrieved pages change between rounds, and it leaves an auditable per-round record of which evidence was consulted. The \emph{iterative reasoning (no memory)} ablation in \cref{tab:main} runs the identical loop but passes an empty bank to every round, so that no notes are carried between rounds; the \emph{single-round} ablation fixes $K{=}1$. Verbatim prompts and the schema are in \cref{app:prompts}.

\section{Results}
\label{sec:results}

\subsection{Setup}
\label{sec:setup}

We evaluate on the MIRAGE~\citep{xiong2024benchmarking} splits: MedQA~\citep{jin2021disease}, 1,273 USMLE questions; MedMCQA~\citep{pal2022medmcqa}, 4,183 questions; PubMedQA~\citep{jin2019pubmedqa}, 500 expert-labeled test questions; and MMLU-Med (MMLU-Medical)~\citep{hendrycks2021measuring}, 1,089 questions across six clinical subdomains. All models we run ourselves are evaluated zero-shot; the option letter is regex-extracted and unparseable outputs count as incorrect. All \MedVRAG{} variants and our own no-RAG baselines use the same prompt template, decoding settings, and extraction procedure. Hardware is 4$\times$NVIDIA A100 80GB. All numbers from our own runs come from a single seeded run (seed 42) with low-temperature decoding (\cref{app:impl}).

\subsection{Main Results}
\label{sec:main}

\begin{table*}[t]
  \centering
  \caption{Accuracy (\%) on the four MIRAGE benchmarks. Rows marked $^\dagger$ are published results copied from the cited papers and use different prompts, decoding, and context; they are shown for reference and are not head-to-head comparable. The \MedVRAG{} block is a controlled sequence: each row toggles exactly one component relative to the row above under identical backbone, prompts, decoding, and evaluation. Bold marks the full system. $^\ast$~HuatuoGPT-o1 reports 63.1\% on MMLU-Pro (Medical), not the MIRAGE MMLU-Med split; the cell is omitted. Med-PaLM~2 wins MedQA and PubMedQA (matched 3-dataset average 80.2 vs.\ our 75.3).}
  \label{tab:main}
  \small
  \setlength{\tabcolsep}{6pt}
  \begin{tabular}{l ccccc}
    \toprule
    \textbf{Model} & \textbf{MedQA} & \textbf{MedMCQA} & \textbf{PubMedQA} & \textbf{MMLU-Med} & \textbf{Avg.} \\
    \midrule
    \multicolumn{6}{l}{\emph{No-RAG baselines}} \\
    GPT-3.5$^\dagger$~\citep{xiong2024benchmarking}              & 60.2 & 62.7 & 78.2 & ---  & --- \\
    GPT-4-base 5-shot$^\dagger$~\citep{nori2023capabilities}    & 86.1 & 73.7 & 77.4 & ---  & --- \\
    Med-PaLM~2$^\dagger$~\citep{singhal2025toward}           & 86.5 & 72.3 & 81.8 & ---  & --- \\
    HuatuoGPT-o1-8B$^\dagger$~\citep{chen2024huatuogpt}      & 72.6 & 60.4 & 79.2 & ---$^\ast$  & --- \\
    Meditron-70B$^\dagger$~\citep{chen2023meditron}         & $\sim$70 & $\sim$66 & 81.6 & 77.6 & 73.8 \\
    Yi-34B (OpenMedLM)$^\dagger$~\citep{maharjan2024openmedlm}   & 72.6 & 68.3 & 77.3 & 81.7 & 75.0 \\
    Qwen3-30B~\citep{yang2025qwen3}                       & 71.5 & 65.1 & 77.7 & 85.0 & 74.8 \\
    Qwen3-8B~\citep{yang2025qwen3}                        & 63.5 & 59.3 & 74.7 & 78.8 & 69.1 \\
    \midrule
    \multicolumn{6}{l}{\emph{Text-only RAG}} \\
    MedRAG + GPT-3.5$^\dagger$~\citep{xiong2024benchmarking}     & 66.6 & 58.0 & 67.4 & 75.5 & 66.9 \\
    MedRAG + GPT-4$^\dagger$~\citep{xiong2024benchmarking}       & 82.8 & 66.7 & 70.6 & 87.2 & 76.8 \\
    Self-BioRAG-7B$^\dagger$~\citep{jeong2024improving}       & 43.6 & 42.1 & ---  & 53.9 & --- \\
    \midrule
    \multicolumn{6}{l}{\emph{\MedVRAG{} controlled sequence (Qwen2.5-VL-32B backbone; one component toggled per row)}} \\
    \quad Backbone only, no RAG                & 71.4 & 62.4 & 73.7 & 83.7 & 72.8 \\
    \quad + text-chunk retrieval (BGE-large), single round   & 74.8 & 66.2 & 73.3 & 86.2 & 75.1 \\
    \quad + page-image retrieval instead of text chunks, single round & 75.8 & 67.0 & 74.9 & 86.8 & 76.1 \\
    \quad + iterative reasoning ($\le 3$ rounds, no memory) & 77.6 & 67.5 & 77.0 & 88.4 & 77.6 \\
    \quad \textbf{+ memory bank (full \MedVRAG{})} & \textbf{79.4} & \textbf{69.2} & \textbf{77.2} & \textbf{88.6} & \textbf{78.6} \\
    \bottomrule
  \end{tabular}
\end{table*}

\Cref{tab:main} reports the main results. Full \MedVRAG{} reaches 78.6\% average accuracy and improves over the same backbone without retrieval by $+5.8$ points, with a positive gain on every dataset (largest on MedQA, $+8.0$, and MedMCQA, $+6.8$). Without retrieval, Qwen2.5-VL-32B scores 72.8\%, compared with 74.8\% for the text-only Qwen3-30B; all retrieval comparisons below use the same VLM backbone throughout. For reference, the published MedRAG\,+\,GPT-4 result is 76.8\% average; even the single-round \MedVRAG{} variant (76.1\%) approaches it with a smaller open-weight backbone, and the full system exceeds it by $+1.8$. We stress that this is a cross-paper comparison with different prompts, decoding, and context windows; the internally controlled evidence is the \MedVRAG{} block of the table, discussed next.

\subsection{Ablation}
\label{sec:ablation}

The \MedVRAG{} block of \cref{tab:main} is an additive sequence in which each row toggles exactly one component relative to the row above. Replacing OCR'd text chunks with page images (rows 2 to 3, both single-round with the Stage-2 filter) yields $+1.0$ on average; letting the VLM iterate for up to three rounds adds $+1.5$; carrying findings across rounds in the memory bank adds a further $+1.0$, for $+5.8$ in total over no retrieval. Per-dataset modality gains are PubMedQA $+1.6$, MedQA $+1.0$, MedMCQA $+0.8$, and MMLU-Med $+0.6$. Two caveats apply. The modality gain is measured against one text retriever (BGE-large-en-v1.5 over 512-token chunks); stronger text pipelines (E5-Mistral, cross-encoder re-rankers, layout-aware parsers such as Nougat) may narrow it. And all deltas come from a single seeded run without confidence intervals, so the two $+1.0$ effects and the $+0.2$ memory gains on PubMedQA and MMLU-Med should be read as suggestive (\cref{sec:limitations}).

\subsection{Rounds, Latency, and the Accuracy--Compute Tradeoff}
\label{sec:iteration}

\begin{figure}[t]
  \centering
  \includegraphics[width=\linewidth]{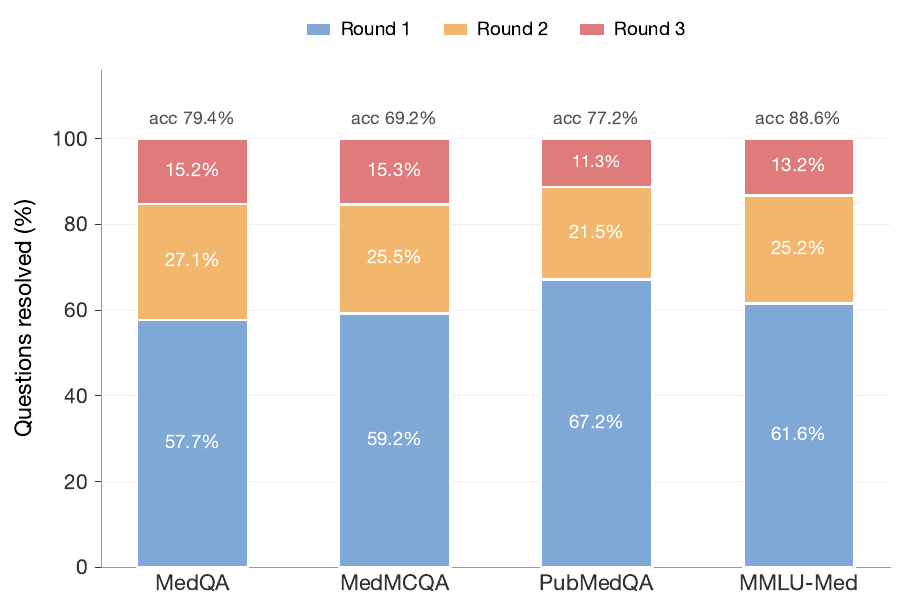}
  \caption{Distribution of reasoning rounds per question, by dataset. Most questions (57.7--67.2\%, mean 61.4\%) are answered in Round~1; 11.3--15.3\% (mean 13.8\%) use all three rounds.}
  \label{fig:iter_dist}
\end{figure}

\begin{figure}[t]
  \centering
  \includegraphics[width=\linewidth]{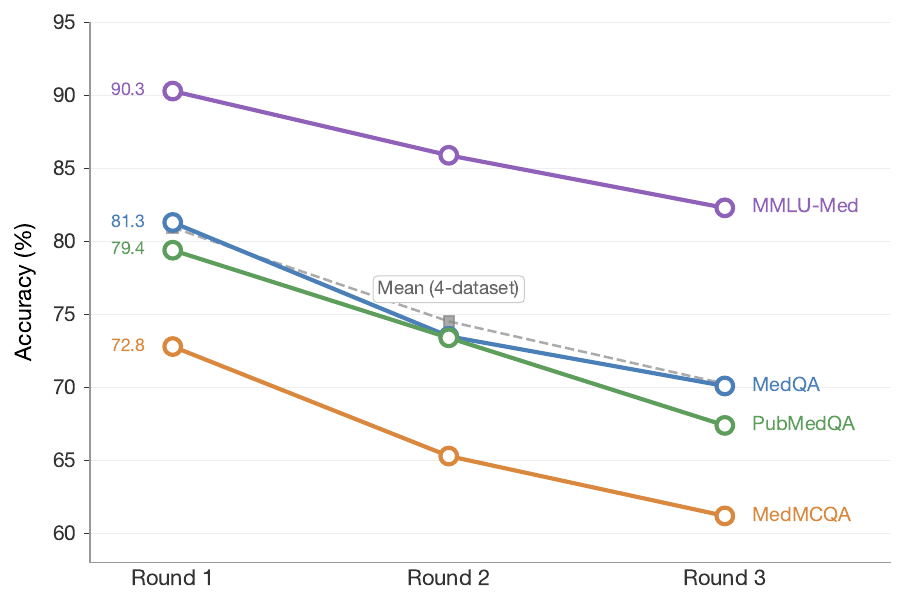}
  \caption{Accuracy conditioned on the number of rounds a question used, per dataset. Later-round questions are harder: the within-dataset weighted average over Round-2 and Round-3 questions is 72.9\%, against 81.0\% for Round-1 questions.}
  \label{fig:iter_acc}
\end{figure}

\begin{table}[t]
  \centering
  \caption{End-to-end latency per query on 4$\times$A100 (wall-clock, mean of 50 MedQA queries). Stage-1 retrieval and the Stage-2 filter are repeated in every round in which the VLM issues a refined query.}
  \label{tab:latency}
  \small
  \setlength{\tabcolsep}{4pt}
  \begin{tabular}{lr}
    \toprule
    \textbf{Stage} & \textbf{Time} \\
    \midrule
    ANN search (CPU)              & $\sim$1.9\,ms \\
    Coarse ranking (CPU)          & $\sim$4.6\,ms \\
    Fine-grained scoring (GPU)    & $\sim$20.9\,ms \\
    LLM filter (Qwen3-30B-A3B, sharded) & $\sim$3.8\,s \\
    VLM reasoning per round   & $\sim$12.1\,s \\
    \midrule
    \textbf{Total (1 round)}                  & $\sim$\textbf{15.9\,s} \\
    \textbf{Total (3 rounds, $3\times$ filter+VLM)} & $\sim$\textbf{47.8\,s} \\
    \bottomrule
  \end{tabular}
\end{table}

Across datasets, 61.4\% of questions are answered in one round, 24.8\% in two, and 13.8\% use all three (\cref{fig:iter_dist}). Questions that stop after Round~1 are easier (mean accuracy 81.0\%); those that continue are harder but still end up largely correct (mean 74.5\% for Round-2 and 70.2\% for Round-3 questions; \cref{fig:iter_acc}). Because the system itself decides which questions continue, these conditional accuracies describe the difficulty of the questions that iterate, not the benefit of iterating; the controlled comparison is the ablation row in \cref{tab:main}, and a forced single-round re-run on the multi-round subset would sharpen it (\cref{sec:limitations}).

\Cref{tab:latency} gives per-stage latency. A single round takes $\sim$15.9\,s and three rounds $\sim$47.8\,s, dominated by the VLM call ($\sim$12.1\,s per round); the embedding retrieval stays under 30\,ms because of the coarse-to-fine index. On MedQA, Round~2 recovers $\sim$7.6\% of the errors made after Round~1 but Round~3 only $\sim$2.5\%, so the marginal accuracy per second drops sharply between the second and third round. Treating the round cap as a deployment knob, a policy of \emph{stop after Round~2 (or earlier if the VLM answers)} would capture roughly $7.6/(7.6{+}2.5)\approx 75\%$ of the errors the three-round pipeline recovers at $\sim$67\% of its worst-case wall time ($\sim$31.7\,s vs.\ $\sim$47.8\,s). We keep the three-round cap throughout for a like-for-like comparison with prior multi-round RAG systems and leave a learned stopping policy conditioned on calibrated confidence to future work.

\subsection{Retrieval Quality and Error Analysis}
\label{sec:errors}

\textbf{Retrieval quality (MedQA).} Relevance labels are produced by a Qwen3-30B-A3B judge and spot-checked by the first author on 100 labels ($\kappa\approx0.81$). Evidence Recall@$N_2{=}100$ is 72.4\% and Recall@$N_1{=}2{,}000$ is 89.4\%. A summary-supported metric, in which the judge sees the cited page's summary rather than its image, reports 85.3\% of correct answers as supported; this is summary-grounded and uses a same-family judge, so it is not a verified visual-grounding rate.

\textbf{Error taxonomy (MedQA).} We assign each error of the full system to one of four mutually exclusive primary causes (single rater): (i)~\emph{retrieval miss} ($\sim$35.9\%), no relevant page in the top-100; (ii)~\emph{filter miss} ($\sim$19.7\%), a relevant page in the top-2,000 dropped by Stage~2; (iii)~\emph{VLM misinterpretation} ($\sim$29.5\%), a relevant page retrieved but reasoned over incorrectly; (iv)~\emph{round drift} ($\sim$14.9\%), a refined query moved off-topic.

\subsection{Analysis}
\label{sec:analysis}

\textbf{Where page images help.} In a manual sample of 100 MedQA questions that full \MedVRAG{} answers correctly and the text-chunk variant does not, we classified the decisive evidence (single rater, one primary cause per question) as figure- or diagram-dependent in 20 cases, table-dependent in 39, and text-whose-layout-matters (section headings, spatial grouping, reference ranges next to prose) in 41. Tables and layout are thus the main beneficiaries, consistent with the intuition that row--column structure and reference ranges linearize poorly into 512-token chunks. The per-dataset modality gains follow the same pattern: MMLU-Med, whose questions are short fact-recall items, shows the smallest gain ($+0.6$), while PubMedQA shows the largest ($+1.6$) despite its single-abstract evidence design; one possible explanation is that page rendering preserves abstract structure (headings, conclusion sentences) that chunking breaks, although the present comparison does not isolate this mechanism.

\textbf{Where iteration helps.} 38.6\% of questions trigger more than one round (\cref{fig:iter_dist}). Inspection shows these are typically multi-hop cases that need mechanism, contraindications, and monitoring schedules from distinct pages, mirroring how a clinician refines a literature search. \Cref{fig:iter_acc} confirms that these questions are intrinsically harder than those answered in one round.

\textbf{When text-only RAG hurts.} \citet{xiong2024benchmarking} report that MedRAG\,+\,GPT-4 \emph{degrades} MedQA-US (84.0$\to$82.8) and MMLU-Med (89.4$\to$87.2) when conflicting text chunks enter the context. Our text-chunk ablation shows the same effect in miniature on PubMedQA (73.7$\to$73.3), a single-source benchmark where extra chunks add noise. \MedVRAG{} does not regress on any dataset, which suggests that page-level evidence plus LLM filtering is more robust to context conflict than naive chunk retrieval, at least relative to the BGE-large baseline used here.

\textbf{Case study (paraphrased).} \emph{Question}: A 45-year-old woman with SLE (joint pain, butterfly rash, elevated anti-dsDNA) is started on hydroxychloroquine. Which side effect is most important to monitor? \emph{Round~1} retrieved 10 hydroxychloroquine pharmacology pages covering indications and dosing but not toxicity monitoring; the VLM issued the refined query \texttt{"hydroxychloroquine retinal toxicity ophthalmologic screening guidelines"}. \emph{Round~2} retrieved 10 ophthalmology pages including an AAO-style monitoring-schedule table; the VLM selected \textbf{retinal toxicity}, citing that page. The full trace with page IDs and image crops will be released with the code.

\section{Limitations}
\label{sec:limitations}

\textbf{Statistical evidence.} All numbers we report for our own runs come from a single seeded run with low-temperature decoding. Without bootstrap confidence intervals, multiple seeds, or paired significance tests, the small deltas ($+1.0$ modality, $+1.0$ memory, $+0.2$ on PubMedQA and MMLU-Med from memory, $+1.8$ cross-paper) are suggestive rather than statistically validated.

\textbf{Baselines and controls.} The text baseline is BGE-large-en-v1.5 over 512-token OCR'd chunks; stronger retrievers, re-rankers, and layout-aware parsers are not evaluated, and neither is a same-backbone reimplementation of an iterative text RAG such as i-MedRAG. The full system also feeds the VLM page summaries alongside page images, whereas the text baseline feeds chunks; a summaries-only control would isolate the contribution of the images themselves. A forced single-round re-run on the multi-round subset would separate the benefit of iterating from the difficulty of the questions that iterate.

\textbf{Scope.} The corpus is PMC-only and the evaluation is multiple-choice. The system is not clinically validated; free-text generation, calibrated uncertainty, harmful-answer analysis, and clinician-in-the-loop review are all required before any deployment. Two design properties are relevant to that path, namely that retrieved evidence stays in its original page-image form and is directly inspectable, and that the memory bank leaves a per-round audit trail; both are necessary but not sufficient.

\section{Conclusion}
\label{sec:conclusion}

\MedVRAG{} retrieves biomedical evidence as page images rather than text chunks and lets a VLM refine its query across rounds while keeping a structured memory of what it has found. On four standard medical QA benchmarks it reaches 78.6\% average accuracy, $+5.8$ over the same backbone without retrieval, and a component-at-a-time ablation reports successive gains from text-chunk retrieval, page-image retrieval, iteration, and memory. The retrieval design scales multi-vector page search to 350K pages at $\sim$30\,ms, and on MedQA most of the errors that iteration recovers are recovered by the second round. The modality and memory effects are small in absolute terms and rest on a single run; establishing them firmly requires confidence intervals, stronger text baselines, and the summaries-only and forced-single-round controls listed above. Beyond that, the natural next steps are extending the corpus to clinical guidelines and textbooks and evaluating on open-ended clinical questions.

\section*{Data and Code Availability}

The page corpus is built from the PMC Open Access Subset (\url{https://www.ncbi.nlm.nih.gov/pmc/tools/openftlist/}). All evaluation benchmarks are public: MedQA~\citep{jin2021disease}, MedMCQA~\citep{pal2022medmcqa}, PubMedQA~\citep{jin2019pubmedqa}, MMLU-Med~\citep{hendrycks2021measuring}. Verbatim prompts, the memory-bank schema, the answer-extraction regex, all decoding settings, and the text-baseline configuration are inlined in \cref{app:prompts,app:impl}. Source code, FAISS indices, page summaries, and reproduction scripts will be released under the MIT license.

\bibliography{references}
\bibliographystyle{icml2026}

\newpage
\appendix
\onecolumn

\section{Prompts and Output Schema}
\label{app:prompts}

\textbf{Stage-2 filter prompt} (Qwen3-30B-A3B, sharded MapReduce; \texttt{target\_k}=25 in map shards, 100 in the reduce call):
\begin{quote}\small\itshape
You are a medical document retrieval expert. Given a medical question and candidate page summaries, select the \texttt{\{target\_k\}} most relevant pages. Question: \texttt{\{question\}}. Candidate page summaries: \texttt{\{summaries\}}. Select the \texttt{\{target\_k\}} most relevant pages by listing their numbers in order of relevance (most relevant first). Output ONLY the page numbers inside \texttt{<selected\_pages>} tags.
\end{quote}

\textbf{VLM reasoning prompt} (Qwen2.5-VL-32B-Instruct):
\begin{quote}\small\itshape
You are a medical QA expert. Answer the multiple-choice question based on the provided document pages. Question: \texttt{\{question\}}. Options: \texttt{\{options\}}. \texttt{\{memory\_section\}}. Retrieved page summaries: \texttt{\{summaries\}}. (The actual page images are also provided for your reference.) Instructions: If you have enough information to answer, output your answer inside \texttt{<answer>} tags with a brief justification. If you need more information, output a refined search query inside \texttt{<query\_update>} tags and summarize your current findings inside \texttt{<notes>} tags. This is iteration \texttt{\{iteration\}/\{max\_iterations\}}. \texttt{\{force\_msg\}}.
\end{quote}
\texttt{\{force\_msg\}} is empty except in the final round, where it instructs the model that it must answer.

\textbf{Memory-bank schema (JSON).} \texttt{\{iteration: int, key\_findings: list[str], reasoning\_history: list[\{iteration, notes\}]\}}. Each \texttt{key\_findings} entry is prefixed \texttt{[Round k]} with the round that produced it; \texttt{reasoning\_history} holds one \texttt{\{iteration, notes\}} pair per non-final round. The serialized object is inserted as \texttt{\{memory\_section\}} in the next round's prompt.

\textbf{Answer extraction (regex).} \texttt{<answer>\textbackslash s*([A-Da-d]|yes|no|maybe)\textbackslash b}; outputs without \texttt{<answer>} tags count as incorrect.

\section{Implementation Details}
\label{app:impl}

\textbf{Models.} ColQwen2.5 ($d{=}128$ after PCA from 768) for page embedding; Qwen2.5-VL-7B for offline page summaries (chosen over the 32B variant for indexing throughput; mean summary length $\sim$120 tokens); Qwen3-30B-A3B for the Stage-2 filter; Qwen2.5-VL-32B-Instruct for reasoning.

\textbf{Retrieval cutoffs.} $N_1{=}2{,}000$ candidates from Stage~1; $N_2{=}100$ pages after the Stage-2 filter; at most 3 rounds (hard cap, enforced regardless of the VLM's output).

\textbf{Decoding.} VLM reasoning $T{=}0.1$, max\_new\_tokens $=2048$; Stage-2 filter $T{=}0$, max\_new\_tokens $=1024$; offline page summaries $T{=}0$. \textbf{Random seed:} 42 for the single reported run. \textbf{Image preprocessing:} Qwen2.5-VL native dynamic-resolution preprocessor with min/max pixels $= 256/1280 \cdot 28^2$. \textbf{Hardware:} 4$\times$NVIDIA A100 80GB.

\textbf{Knowledge-base construction.} The PMC Open Access Subset is restricted to articles published between 2000 and 2024. Each PDF is rendered page-by-page at 300\,DPI to PNG. Near-duplicate pages are removed by ColQwen2.5 cosine-similarity thresholding at $0.97$. The final corpus contains $\sim$350K pages from $\sim$18K articles. The FAISS index over PCA-reduced patch vectors holds $\sim$35.9M patch-level entries (avg.\ $\sim$103 patches per page). Leakage check: articles overlapping with MedQA/MedMCQA/PubMedQA source articles are excluded by DOI/PMID; for MMLU-Med (no source DOIs) we additionally verified zero verbatim question-text matches in retained pages.

\textbf{Text-chunk retrieval baseline} (the ``+ text-chunk retrieval'' row of \cref{tab:main}). Pages are OCR'd, chunked into 512-token segments with 64-token stride, embedded with BGE-large-en-v1.5, and indexed in a separate FAISS index. Stage~1 returns the top-2{,}000 chunks by cosine similarity; Stage~2 (the same Qwen3-30B-A3B sharded MapReduce filter) selects $N_2{=}100$ chunks; the VLM then receives the chunks as text rather than page images. All other settings (single round, prompts, decoding) are identical to the page-image single-round variant; the two rows therefore differ in retrieval modality and in whether the VLM also receives page summaries (\cref{sec:limitations}).

\textbf{Dataset details.} MedQA: 1{,}273 USMLE 4-option questions covering clinical medicine, pharmacology, and pathology. MedMCQA: 4{,}183 validation questions from AIIMS/NEET across 21 medical subjects. PubMedQA: 500 expert-labeled test questions following the MIRAGE split (yes/no/maybe). MMLU-Med: 1{,}089 questions across six subdomains (Clinical Knowledge, Medical Genetics, Anatomy, Professional Medicine, College Biology, College Medicine).

\textbf{Per-dataset round distribution} (numbers behind \cref{fig:iter_dist}):
MedQA $\{$R1 57.7\%, R2 27.1\%, R3 15.2\%$\}$;
MedMCQA $\{$R1 59.2\%, R2 25.5\%, R3 15.3\%$\}$;
PubMedQA $\{$R1 67.2\%, R2 21.5\%, R3 11.3\%$\}$;
MMLU-Med $\{$R1 61.6\%, R2 25.2\%, R3 13.2\%$\}$.

\textbf{Per-round accuracy} (numbers behind \cref{fig:iter_acc}, equal-weight 4-dataset mean): R1 81.0\%, R2 74.5\%, R3 70.2\%. The within-dataset weighted average over R2+R3 questions is 72.9\%.

\end{document}